\documentclass[journal]{IEEEtran}
\usepackage{graphicx}
\usepackage{amsmath,amssymb,bm} % define this before the line numbering.
\usepackage{color}
\usepackage{caption}
\usepackage{subfig}
\usepackage{booktabs}
\usepackage{cite}
\usepackage{soul}
\usepackage{algorithm}
\usepackage{algpseudocode}
\usepackage{stackengine}
\DeclareMathOperator*{\argmin}{arg\,min}
\setstackEOL{\\}

\begin{document}

\newcommand{\tabincell}[2]{\begin{tabular}{@{}#1@{}}#2\end{tabular}}
\newcommand{\kun}[1]{\textcolor{black}{#1}}
\def\BState{\State\hskip-\ALG@thistlm}

%Kun Hu, Zhiyong Wang, Wei Wang, Kaylena Ehgoetz Martens, Liang Wang, Tieniu Tan, Simon Lewis, David Dagan Feng
\title{Multi-level Adversarial Spatio-temporal Learning for Footstep Pressure based FoG Detection}
\author{Kun~Hu,~\IEEEmembership{Member,~IEEE,}
        Shaohui~Mei,~\IEEEmembership{Member,~IEEE,}
        Wei~Wang,~\IEEEmembership{Member,~IEEE,}
        Kaylena~A.~Ehgoetz~Martens,\\
        Liang~Wang,~\IEEEmembership{Fellow,~IEEE,}
        Simon~J.~G.~Lewis,
        David~D.~Feng,~\IEEEmembership{Fellow,~IEEE}
        and~Zhiyong~Wang,~\IEEEmembership{Member,~IEEE}
        % <-this % stops a space
\thanks{This work was supported in part by the Australian Research Council (ARC) under
Grant DP210102674 and DP160103675, in part by the NHMRC-ARC Dementia Fellowship
under Grant 1110414, in part by the National Health and Medical Research
Council (NHMRC) of Australia Program Grant under Grant 1037746, in part by the Dementia Research Team under Grant 1095127, in part by the
NeuroSleepCentre of Research Excellence under Grant 1060992, in part by the ARC Centre of Excellence in Cognition and Its Disorders Memory Program under Grant CE110001021, in part by the Sydney Research Excellence Initiative (SREI) 2020 of the University of Sydney, in part by the Natural Science Foundation of China under Grant 61420106015, in part by the Parkinson Canada, and in part by the Brain and Mind Centre Early Career Research Development Grant.}% <-this % stops a space
\thanks{K. Hu, Z. Wang, and D. D. Feng are with the School of Computer Science, The University of Sydney, NSW 2006,
Australia (E-mail: kuhu6123@uni.sydney.edu.au; zhiyong.wang@sydney.edu.au; dagan.feng@sydney.edu.au).}% <-this % stops a space
\thanks{S. Mei is with the School of Electronics and Information, Northwestern Polytechnical University, Xi'an, 710072, China (E-mail: meish@nwpu.edu.cn).}% <-this % stops a space
\thanks{W. Wang, and L. Wang are with the Center for Research on Intelligent Perception and Computing (CRIPAC), National Laboratory of Pattern Recognition (NLPR), Institute of Automation Chinese Academy of Sciences (CASIA) Beijing 100190, China, and University of Chinese Academy of Sciences (UCAS). L. Wang is also with the Center for Excellence in Brain Science and Intelligence Technology (CEBSIT), Institute of Automation Chinese Academy of Sciences (CASIA) (E-mail: wangwei@nlpr.ia.ac.cn;wangliang@nlpr.ia.ac.cn).}% <-this % stops a space
\thanks{K. A. Ehgoetz Martens is with University of Waterloo, Canada (E-mail: kaehgoet@uwaterloo.ca).}% <-this % stops a space
\thanks{S. J. G. Lewis is with the ForeFront Parkinson’s Disease Research Clinic, Brain and Mind Centre, The University of Sydney, Sydney, NSW 2050, Australia (E-mail:  simon.lewis@sydney.edu.au).}% <-this % stops a space
}

\maketitle

\begin{abstract}

Freezing of gait (FoG) is one of the most common symptoms of Parkinson's disease, which is a neurodegenerative disorder of the central nervous system impacting millions of people around the world. To address the pressing need to improve the quality of treatment for FoG, devising a computer-aided detection and quantification tool for FoG has been increasingly important. As a non-invasive technique for collecting motion patterns, the footstep pressure sequences obtained from pressure sensitive gait mats provide a great opportunity for evaluating FoG in the clinic and potentially in the home environment. In this study, FoG detection is formulated as a sequential modelling task and a novel deep learning architecture, namely Adversarial Spatio-temporal Network (ASTN), is proposed to learn FoG patterns across multiple levels. A novel adversarial training scheme is introduced with a multi-level subject discriminator to obtain subject-independent FoG representations, which helps to reduce the over-fitting risk due to the high inter-subject variance. As a result, robust FoG detection can be achieved for unseen subjects. The proposed scheme also sheds light on improving subject-level clinical studies from other scenarios as it can be integrated with many existing deep architectures. To the best of our knowledge, this is one of the first studies of footstep pressure-based FoG detection and the approach of utilizing ASTN is the first deep neural network architecture in pursuit of subject-independent representations. Experimental results on 393 trials collected from 21 subjects demonstrate encouraging performance of the proposed ASTN for FoG detection with an AUC 0.85.

\end{abstract}


\section{Introduction}
\IEEEPARstart{M}{illions} of people around the world are impacted by the 
Parkinson's disease (PD) which is a neurodegenerative disease predominantly characterized by its effects on the motor system \cite{dorsey2018global}. Freezing of gait (FoG) is one of the most common PD symptoms, identified by its sudden and brief episodes of cessation of movement despite the intention of a patient to keep walking \cite{hely2008sydney,macht2007predictors}. With the progression of the disease, FoG happens more and more frequently, becoming a major risk factor for falls \cite{bloem2004falls,LEWIS2009333} and eventually affects the mobility, independence and quality of life of a PD patient and their family \cite{schaafsma2003characterization}.
Accurate detection and quantification of FoG are of great importance in clinical practice and could be used for assessing the impact of a treatment \cite{donovan2011laserlight}.
However, current manual annotation of FoG events relies heavily on subjective scoring by well-trained experts, which is extremely time-consuming and variable.
Therefore, computer aided intelligent solutions are required for timely, accurate and objective FoG detection.

\begin{figure}[!htb]
\centering
\includegraphics[width=0.49\textwidth]{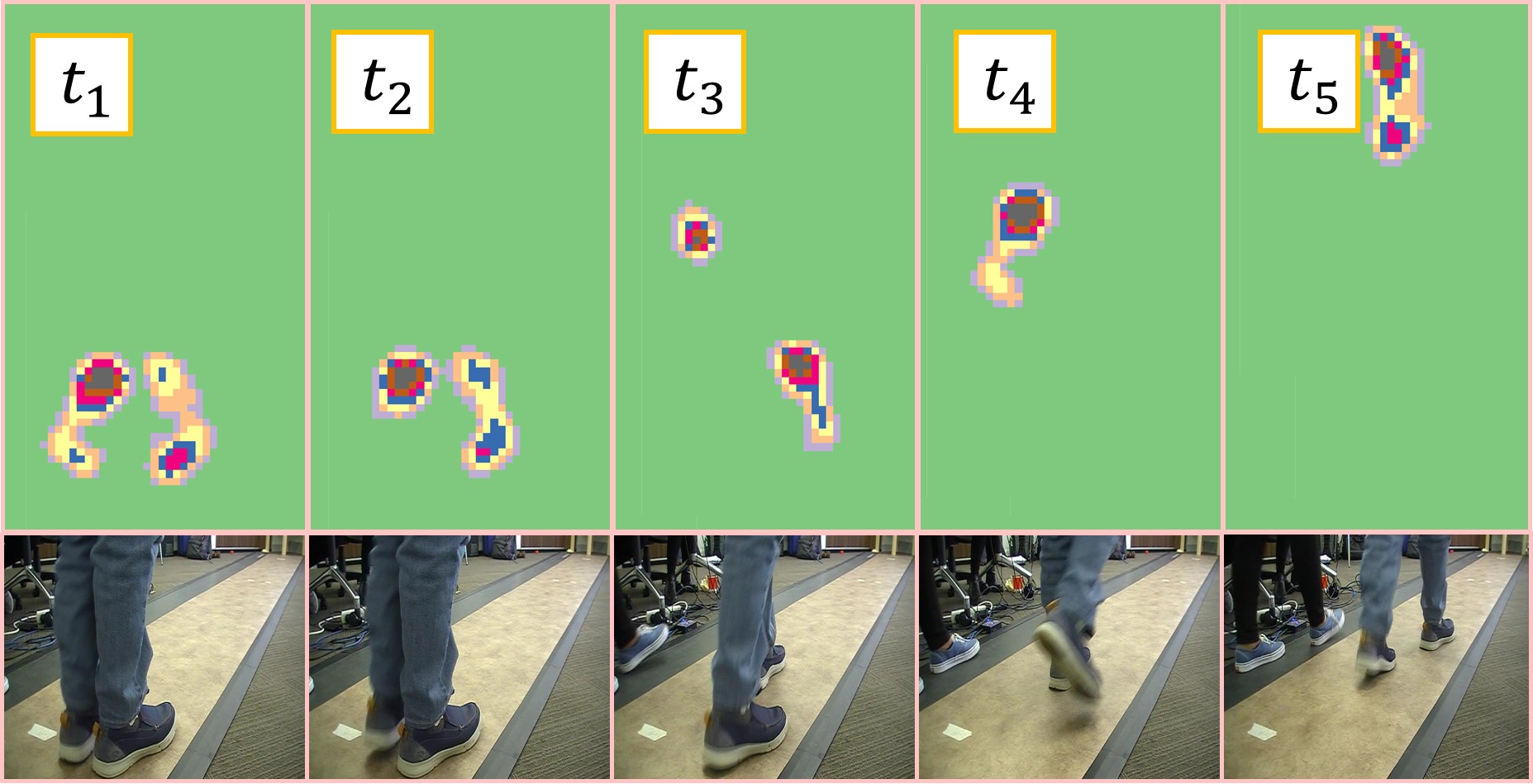}
\caption{Illustration of a footstep pressure sequence collected from a pressure mat during a clinical assessment. Continuous pressure heat maps and corresponding video recording frames are shown to help understand the data acquired from a trial.}
\label{fig:sample_demo}
\end{figure}

In general, existing studies focusing on computer-aided FoG detection are based on two major types of approaches: wearable sensor-based methods \cite{bachlin2010wearable,tripoliti2013automatic,ravi2017deep,rodriguez2017home,prateek2018modeling,thorp2018monitoring} and vision-based methods \cite{k8737782,hu2019graph}. 
Nonetheless, these methods need additional efforts to be deployed in complex real environments, which would otherwise limit their general applicability. 
As a non-obtrusive and less environment-dependent protocol for collecting motion related patterns, the footstep pressure sequence obtained from a pressure mat provides a convenient and promising opportunity for accurate FoG characterization and detection for the purpose of both clinical assessments and daily home care.
%A pioneer footstep pressure study based on deep learning solving the biometric verification problem by treating the pressure sequence as a 2-dimensional image series\cite{costilla2019analysis}. Similarly, 
As illustrated in Fig. \ref{fig:sample_demo}, the footstep pressure sequence collected in a clinical FoG assessment is organized in temporal order and pressure heat maps are drawn for a number of selected temporal indices.
However, the modality of footstep pressure has seldom been investigated for FoG detection, thus appropriate methods are currently required to fully explore the potential of such sequential data for accurate FoG detection.

There has been ground-breaking success of deep learning techniques for many tasks such as object detection and human action recognition \cite{lecun2015deep,qian2021diversifying}, which provides a great opportunity for developing deep learning based methods to accurately characterize FoG patterns from footstep pressure sequences. 
In this scheme, pressure mat based FoG detection can be formulated as a sequence to sequence task where a pressure mat sequence is mapped into a FoG output sequence through temporal neural networks such as recurrent neural networks (RNNs). However, considering that the variations among all subjects (inter-subject variance) could be higher than the variance between FoG and non-FoG events (inter-class variance), there could be the potential issues to directly adopt these deep architectures as they were devised for general tasks often involving high inter-class variance. 
For subject-level clinical applications including FoG detection, such large inter-subject variance of the feature vector (representation) could negatively impact the model performance when applying the trained model to unseen subjects due to the potential of over-fitting risk on subject specified patterns. 
Hence, it is critical to devise proper mechanisms to reduce the inter-subject variance in pursuit of deriving the subject-independent FoG representation.

In this study, a novel end-to-end deep architecture, namely Adversarial Spatio-Temporal Network (ASTN), is proposed to address the FoG detection problem by utilizing the footstep pressure sequence at three levels. 
As FoG exhibits both the short-term intrinsic property and the long-term dynamic characteristic, ASTN introduces temporal convolutions and RNN cells together to formulate multi-level temporal FoG patterns from the sequentially organized footstep pressure data. 
In addition, by introducing an adversarial training scheme with a multi-level discriminator, ASTN is able to reduce the inter-subject variance and formulate subject-independent FoG representations. With the proposed mechanisms, it is expected that the subject-independent representation can be obtained for effective FoG detection.

In summary, the major contributions of this paper are four-fold:
\begin{itemize}
  \item We propose pressure mat based FoG detection as a sequential modelling task, which is one of the first studies utilizing footstep pressure data for FoG detection.
  \item A novel end-to-end deep architecture ASTN with an adversarial training scheme is proposed to characterize subject-independent FoG representations by reducing the inter-subject variation.
  \item The proposed architecture sheds light on improving other subject-level clinical modelling tasks, as it is available to integrate the adversarial training scheme with many deep learning architectures. 
  \item A large footstep pressure sequence dataset was created during clinical assessments of 21 subjects to evaluate the effectiveness of our proposed methods.
\end{itemize}

The rest of the paper is organized as follows.
Section \ref{sec:related} reviews the related works on FoG detection, footstep pressure data analysis, and spatio-temporal deep learning techniques.
Section \ref{sec:method} introduces the details of our proposed method.
Section \ref{sec:evaluation} presents comprehensive experimental results to evaluate the effectiveness of our proposed footstep pressure based FoG detection method.
Lastly, Section \ref{sec:conclusion} concludes our study with discussions on our future work.

\section{Related Work}
\label{sec:related}

In this section, the related studies are reviewed from three aspects. Firstly, existing freezing of gait detection methods are discussed in terms of traditional wearable sensor systems and vision cameras. Secondly, as our method utilized footstep pressure sequence data, key methods of existing research on footstep pressure data analysis are reviewed. 
Lastly, considering that footstep pressure sequences contain both spatial and temporal patterns, we focus on the potential deep learning based techniques to deal with the combined modalities by the spatio-temporal data. 

\subsection{Freezing of Gait Detection for Parkinson's Disease}

In general, existing studies related to automatic FoG detection are based on two major approaches: wearable sensor-based methods and vision-based methods. 
For traditional sensor-based FoG detection methods \cite{thorp2018monitoring,bachlin2010wearable,tripoliti2013automatic,ravi2017deep,rodriguez2017home,prateek2018modeling}, sensor systems are designed similar to general gait analysis methods (e.g. \cite{zou2017robust}) including accelerometers, gyroscopes or by pairing these two sensors in an inertial measurement unit to capture temporal signals. Machine learning methods such as support vector machine, random forest, and deep learning are further utilized to analyze and detect the FoG patterns. 
Note that an ideal sensor system is expected to be simple and unobtrusive for deploying for the goal of monitoring FoG symptoms during clinical assessments and home care,
whilst current sensor-based studies require that sensors be placed on a number of anatomical locations where FoG patterns can potentially be recorded.

Recently, to address the limitations of these traditional sensor-based methods, vision-based methods for FoG detection \cite{k8737782,hu2019graph} have been proposed by taking advantage of its natural and unobtrusive characteristics \cite{zhang2017gii}. These studies collect videos from clinical assessments during which the subject follows instructions to participate in a timed up and go (TUG)  gait test. Such TUG tasks assess the functional mobility of a person in a standardized fashion  \cite{shumway2000predicting}. 
Deep learning architectures are devised to characterize and detect fine-grained FoG patterns in these studies due to their great success in solving many vision related tasks. 
Although these studies have achieved promising results, there may be potential issues when applying them in practical environments which are generally more complex compared with the experimental protocols used during clinical data collection.
For example, these studies work for cameras with fixed placements and require well-lit conditions, without shadows and occlusions. %Hence, it is possible that the vision-based methods fail to capture efficient FoG patterns for general scenarios.

Therefore, footstep pressure sequence data obtained from a pressure mat is an attractive method for monitoring FoG events in both clinical and home care protocols. Moreover, the pressure data provides the chance for multimodal learning \cite{wang2021survey} with data collected using other modalities such as videos in pursuit of robust FoG detection. 

\subsection{Footstep Pressure Data Analysis}

In addition to analyzing the gait of patients for clinical purposes \cite{lee2017gait,guo2017classification}, %FoG detection, 
footstep pressure (sequence) data has been used for machine learning applications such as biometrics \cite{costilla2016temporal,vera2012comparative,middleton2005floor}.
Most of them follow a traditional pattern recognition pipeline: extracting hand-crafted features first and then feeding these features into machine learning models such as generalized linear models (GLM), support vector machines (SVM), ensemble learning methods, and hidden Markov models (HMM) to produce prediction outputs. 
However, few evidences demonstrate that these existing hand-crafted features are well associated with FoG detection. %to generate accurate predictions. %Furthermore, it is complex and time-consuming to design and verify the effectiveness of new hand-crafted features for FoG detection since understanding the footstep pressure sequence is not straightforward for clinicians.

Very recently, deep learning has been introduced to deal with footstep pressure sequence data. By treating a footstep pressure sequence as 2-dimensional image series, a two-stream architecture is devised to process the spatial and the temporal streams of a footstep pressure sequence \cite{costilla2019analysis}.
The pipeline is similar to general deep learning models taking image inputs, whilst the spatial information of a footstep sequence at a specified temporal index contains a one-channel heat map instead of 3 channels as in general RGB images. 
A simplified version of the ResNet-50 is adopted for each stream to learn spatial and temporal patterns, respectively. 
Note that this study treats the temporal dimension of an input sequence as an image channel, thus the approach may not characterize temporal patterns well. 

\subsection{Deep Learning for Spatio-temporal Data}
A footstep pressure sequence contains both spatial and temporal patterns, which can be viewed as a 2-dimensional grey image sequence or a 3-dimensional vision cube. 
Deep learning architectures devised for the spatio-temporal data have demonstrated great success across a wide range of tasks including action recognition and object detection and tracking. %These architectures are reviewed as follows. 
Initially to process 2-dimensional spatial sequences, single-stream \cite{karpathy2014large} and two-stream methods \cite{simonyan2014two} are proposed. 
The single stream based methods utilize pre-trained 2-dimensional convolution filters frame by frame and a number of temporal fusion strategies are being further investigated. The two-stream based methods take advantage of the appearance and the optical flow features obtained by 2-dimensional convolutions to form spatio-temporal representations.

Based on these pioneering studies, three major types of deep learning methods are currently utilized to process spatio-temporal inputs: convolution neural network (CNN) based methods, RNN based methods and two-stream based methods.
The first type in general extends the 2-dimensional CNN architecture to a 3-dimensional counterpart by which the convolution filters are extended to filter 3-dimensional inputs,
such as C3D \cite{ji20133d,tran2015learning}, P3D \cite{qiu2017learning} and I3D \cite{xie2018rethinking}.
By considering the input as a 2-dimensional spatial sequence, the second type aims to model the temporal structure with RNNs \cite{donahue2015long,xingjian2015convolutional,zhang2018spatial}, which have been widely used to model sequential patterns. 
In particular, LSTM (Long Short-Term Memory) and GRU (Gated Recurrent Unit) networks are proposed to address the gradient vanishing and exploding issues of the original RNN by using gate mechanism \cite{hochreiter1997long,cho2014learning}. GRU involves less computations than LSTM while achieving comparable performance to LSTM, thus improving the efficiency of RNNs. Moreover, GRU has demonstrated better performance on smaller datasets \cite{chung2014empirical}. 
The last type which is based on the pioneering two-stream approach represents video content with both appearance and motion features \cite{feichtenhofer2016convolutional}. 

The methods mentioned above mainly address the general spatio-temporal related problems, which involve significant inter-class variations. However, intra-class variations could be dominant in many applications such as FoG detection.
%However, the variation between subjects could be dominant instead of the variation between FoG and non-FoG cases.
Hence, fine-grained architectures are proposed for the applications with subtle inter-class variations. In summary, there are two major approaches of these architectures.
Firstly, the patch-based methods (e.g. \cite{wang2018videos,wu2018deep} and our recent one \cite{hu2019graph}) have been proposed to model the subtle variations by utilizing regional or patch based patterns to characterize inter-class variations. 
The patch-based approaches are able to further analyze the localized key patterns. 
However, for footstep pressure sequences, it is not straightforward to identify such patches, considering the data modality as shown in Fig.\ref{fig:sample_demo}. 
Secondly, a bilinear pooling approach is proposed to eliminate the need for region related prior knowledge \cite{girdhar2017attentional}.
Nonetheless, the bilinear pooling increases the model complexity, whilst footstep pressure sequence data is sensitive to complex architectures and the FoG detection performance could be compromised by directly following this approach.
%FoG related variation from the footstep pressure sequence collected during the clinical assessment. 

\begin{figure*}[!htb]
\centering
\includegraphics[width=1\textwidth]{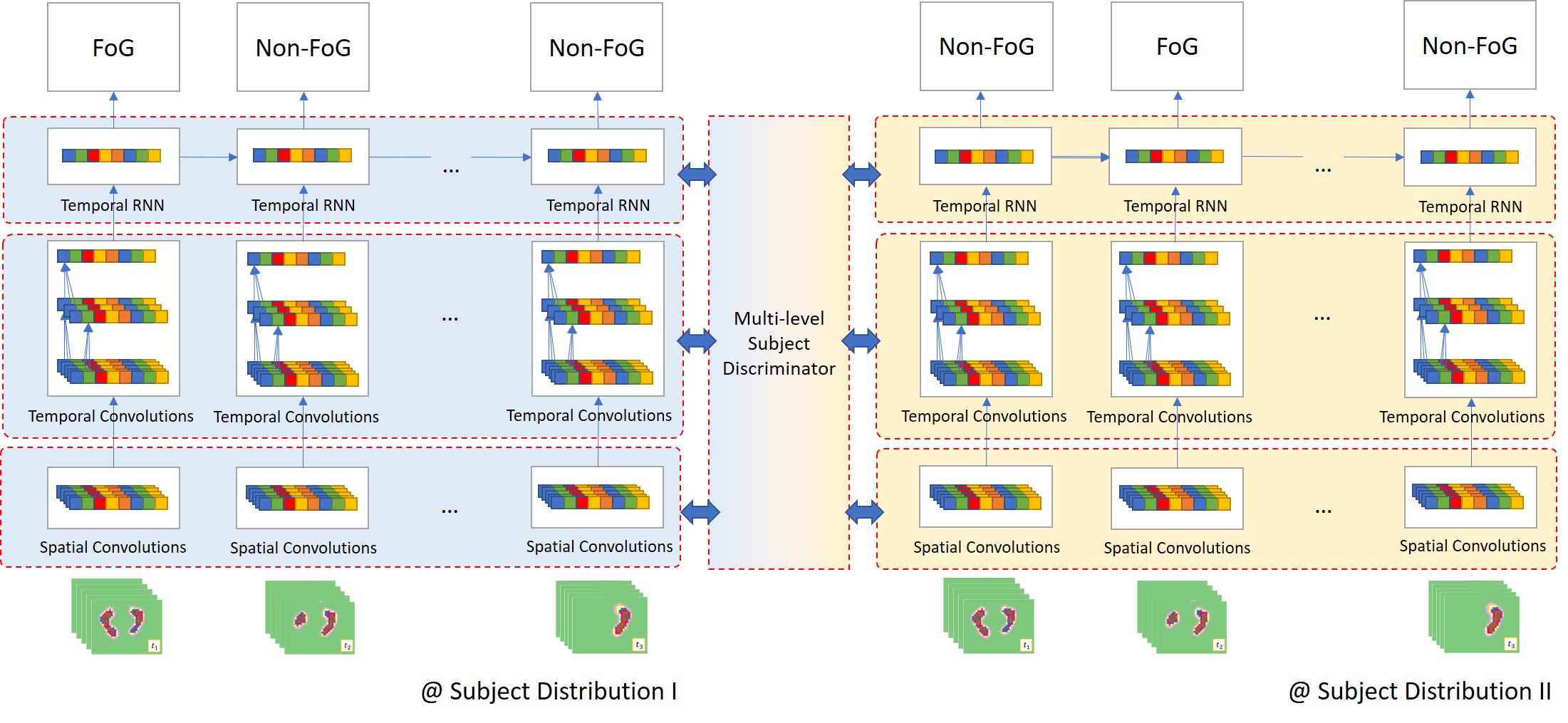}
\caption{Illustration of the proposed ASTN architecture for FoG detection. There are two key components.
Firstly, ASTN involves three levels of spatio-temporal representation: spatial patterns, intrinsic FoG patterns and dynamic temporal patterns obtained from AlexNet, temporal convolutions and GRU, respectively. 
Secondly, an adversarial training scheme is proposed to alleviate the potential gap among subjects by introducing a multi-level subject discriminator. 
 }
\label{fig:architecture}
\end{figure*}

\section{Proposed Method}
\label{sec:method}

As shown in Fig. \ref{fig:architecture}, our novel adversarial spatial-temporal network (ASTN) consists of two key components: a three-level spatial-temporal feature extractor and a multi-level subject discriminator. % is proposed to formulate the subject-independent FoG representation.
Firstly, for each trial, three-level spatio-temporal representations are obtained, including spatial representation, intrinsic FoG representation and dynamic temporal representation.
Secondly, a multi-level subject discriminator is introduced with an adversarial training scheme to determine whether the trial representations contain any patterns associated with subject specific distributions.
In this section, we first introduce footstep pressure sequence data and its spatial and temporal representations, FoG classifier, and multi-level subject discriminator, followed by adversarial training scheme. 
%the details of each component of the ASTN are discussed.

\subsection{Footstep Pressure Sequence Data}
Footstep pressure sequences were obtained during the clinical assessment for each subject with a number of trials. Pressure data of each trial was collected from a set of sensors at a specified sampling rate. In total, $W\times H$ sensors recorded the pressure levels of each frame (i.e., sample) along the temporal indices. %, which is an integer between 0 and 10. 
By taking the locations of these sensors into consideration, the signals collected from them can be viewed as a 2-dimensional pressure heat map for each frame. 
A matrix $\mathbf{X}_{t, p}^{m, n} \in \mathbb{R}^{W \times H}$ is used to represent this pressure map, where $t$, $p$, $m$ and $n$ indicates temporal index (second level), frame index, subject index and trial index, respectively. 
%Comparing with the general 3-channel RGB image, 
The matrix can be viewed as a grey scale image.
Hence, a footstep pressure map sequence $\{\mathbf{X}_{t, p}^{m, n}, t = 1, ... , T^{m, n}, p = 1,...,P\}$ is collected to illustrate a trail of the clinical assessment, where $T^{m, n}$ indicates the total length of the trial $n$ of subject $m$ and $P$ indicates the sampling rate per second. 
Note that the sequence contains both spatial and temporal patterns.
In addition, denote $y_{t}^{m, n} \in \{0, 1\}$ as the binary response to be associated with each temporal index of the spatial pressure map sequence, which indicates whether FoG occurs within the temporal window of index $t$ or not (i.e., 1 for FoG event if at least one frame is annotated as FoG or otherwise 0 for non-FoG event). 

\subsection{Spatial Representation}
Given a temporal index $t$ and a frame index $p$, a footstep pressure map is derived from a grid of sensors. %Besides the patterns of the pressure level included by the pressure map, signal location relations are also included in line with the placement of the sensors. 
%It is natural to address the two aspects of the footstep pressure maps simultaneously for better understanding and analyzing FoG patterns.
Intuitively, a pressure heat map can be interpreted as a grey image and deep learning techniques such as CNNs can be used for learning representation of pressure data based FoG patterns. 

Various CNN based methods have been proposed for image representation through a series of convolution filters, pooling layers and activation functions. 
In general, the recent studies of CNNs tend to construct deeper architectures, which are able to learn complex representations for improved classification or recognition performance.
%For example, AlexNex, ResNet and DenseNet achieved promising performance. 
However, footstep pressure maps are quite different from general images: only a small area of sensors precept pressure levels higher than zero and a limited range of pressure levels for each position are obtained to produce a pressure map. 
In addition, for image related tasks, transfer learning has been widely utilized to fine-tune pre-trained deep networks, whilst it is required to train from scratch for footstep pressure maps due to its unique modality. 

Therefore, instead of training a sufficiently deep neural network such as ResNet or DenseNet, AlexNet is utilized to learn spatial patterns from footstep pressure heat maps. Note that the ReLU activation functions of the original AlexNet are altered to the leaky version of ReLU to help transfer gradient across layers during the forward and backward propagations without a hitch:
\begin{equation}
\mathrm{LeakyReLU}(x) = \left\{
             \begin{array}{lr}
             x & if x \geqslant 0,  \\
             wx & otherwise,  
             \end{array}
\right.
\end{equation}
where $w$ is the negative slope controlling the extent of layer patterns to be kept when $x<0$. 
By denoting the function of obtaining spatial representation as $g(x)$, for each spatial input $\mathbf{X}_{t, p}^{m, n}$, the spatial representation can be written as: 
\begin{equation}
\mathbf{s}_{t, p}^{m, n} = g(\mathbf{X}_{t, p}^{m, n}),
\end{equation}
where $\mathbf{s}_{t, p}^{m, n} \in \mathbb{R}^{S}$ and $S$ indicates the dimension of the spatial representation.

\subsection{Temporal Representation}

As FoG is a kind of movement related symptoms, temporal patterns are critical as well for precise characterization and identification of FoG events. To characterize temporal patterns, in this study, a two-level strategy was adopted. Firstly, the short-term intrinsic FoG patterns were learned to characterize common FoG characteristics, which were independent of each sequence. 
Secondly, the global sequential FoG patterns were further learned to characterize th FoG dynamics of a particular sequence.

To capture the intrinsic FoG patterns, non-overlapped sliding temporal windows, of which the duration was one second, were applied to each footstep pressure sequence. Within each temporal window, a sub-sequence of $P$ frames was derived and further applied with temporal convolutions. 
Since the convolution filters are invariant across all sub-sequences obtained for each temporal window, they were expected to learn intrinsic FoG patterns independent of each sequence. 
In detail, denote
\begin{equation}
\mathbf{s}_{t}^{m, n} = (\mathbf{s}_{t, 1}^{m, n},...,\mathbf{s}_{t, P}^{m, n})^{T} \in \mathbb{R}^{P\times S},
\end{equation}
to represent the spatial representations within the $t$-th temporal window. 
Note that $\mathbf{s}_{t}^{m, n}$ can be viewed as a 1-dimensional temporal sequence of $S$ channels. 
Hence, a series of temporal convolution filters, pooling layers and activation functions can be applied to compute $\mathbf{s}_{t}^{m, n}$. 
By defining the computation as $h(x)$, the intrinsic temporal representation of each second can be summarized as: 
\begin{equation}
\dot{\mathbf{g}}_{t}^{m, n}=h(\mathbf{s}_{t}^{m, n}),
\end{equation}
where $\dot{\mathbf{g}}_{t}^{m, n} \in \mathbb{R}^{H_{1}}$ and $H_{1}$ is the dimension of the intrinsic temporal representation.

In general, FoG patterns vary in these sequences in terms of severity and duration, and it is necessary to characterize such sequence-level dynamic patterns for accurate FoG detection.  
%To further characterize FoG patterns related to a specified subject and trial, which includes the correlations over the entire temporal series, recurrent neural network is adopted. 
To this end, RNNs are employed to model such sequential relations by memorizing proper historical states. 
Recently, gated mechanism introduced to RNNs such as the LSTM and the GRU architecture alleviates the gradient vanishing and exploding issues by controlling the extent to keep patterns from the historical states and to update patterns from the current state. 
In this study, GRU was utilized to characterize the dynamic temporal patterns due to its lower model complexity compared with LSTM. 
The computations of GRU take the intrinsic temporal representations of an entire trial $\{\dot{\mathbf{g}}_{t}^{m, n}, t = 1, ..., T^{m, n}\}$ as the input. In detail, the dynamic temporal patterns can be computed within GRU cells as: 

\begin{equation}
\mathbf{z}_{t}^{m, n}:=\sigma (\mathbf{W}_{xz}\dot{\mathbf{g}}_{t}^{m, n}+\mathbf{W}_{hz}\ddot{\mathbf{g}}_{t-1}^{m, n} + \mathbf{b}_{z}),
\label{equ:graphgrucell_1}
\end{equation}
\begin{equation}
\mathbf{r}_{t}^{m, n}:=\sigma(\mathbf{W}_{xr}\dot{\mathbf{g}}_{t}^{m, n}+\mathbf{W}_{hr}\ddot{\mathbf{g}}_{t-1}^{m, n} + \mathbf{b}_{r}),
\label{equ:graphgrucell_2}
\end{equation}
\begin{equation}
\mathbf{h}_{t}^{m, n}:=tanh(\mathbf{W}_{xh}\dot{\mathbf{g}}_{t}^{m, n}+\mathbf{r}_{t}^{m, n}\odot \mathbf{W}_{hh}\ddot{\mathbf{g}}_{t-1}^{m, n} + \mathbf{b}_{h}),
\label{equ:graphgrucell_3}
\end{equation}
\begin{equation}
\ddot{\mathbf{g}}_{t}^{m, n}:=(1-\mathbf{z}_{t}^{m, n})\mathbf{h}_{t}^{m, n}+\mathbf{z}_{t}^{m, n}\ddot{\mathbf{g}}_{t-1}^{m, n},
\label{equ:graphgrucell_4}
\end{equation}
where $\mathbf{z}_{t}^{m, n}$ and $\mathbf{r}_{t}^{m, n}$ are the reset gate and update gate, respectively; $\mathbf{b}_{z}$, $\mathbf{b}_{r}$ and $\mathbf{b}_{h}$ are the bias terms; $\mathbf{h}_{t}^{m, n}$ is the hidden state of the cell $t$; $\ddot{\mathbf{g}}_{t}^{m, n} \in \mathbb{R}^{H_{2}}$ is the dynamic temporal representation, of which the dimension is $H_2$.
By introducing the gated mechanism, GRU was able to keep the track of proper dependencies between footstep pressure maps across the sequence efficiently. 

Note that the computations of Eq. (\ref{equ:graphgrucell_1}) - (\ref{equ:graphgrucell_4}) illustrate a forward GRU architecture, of which each state only depends on its predecessors. Therefore, the forward architecture can be naturally deployed for the online FoG prediction purpose. 
Furthermore, additionally involving a backward GRU architecture to summarize future patterns and concatenating with the patterns of historical states, a bi-directional GRU is adopted for fully modelling sequential FoG patterns. 

\subsection{FoG Classifier}
The spatial-temporal representation including $\mathbf{s}_{t}^{m, n}$, $\dot{\mathbf{g}}_{t}^{m, n}$ and $\ddot{\mathbf{g}}_{t}^{m, n}$ has been derived in line with the above discussions. 
Denote a function $G$ to summarize all computations to generate this spatio-temporal representation:
\begin{equation}
\{\mathbf{s}_{t}^{m, n}, \dot{\mathbf{g}}_{t}^{m, n}, \ddot{\mathbf{g}}_{t}^{m, n}\} = G(\{\mathbf{X}_{t, p}^{m, n}\}).
\end{equation}

Note that the dynamic temporal pattern $\ddot{\mathbf{g}}_{t}^{m, n}$ is based on the computations of the spatial pattern $\mathbf{s}_{t}^{m, n}$ and the intrinsic pattern $\dot{\mathbf{g}}_{t}^{m, n}$, and thus contains the other two levels of patterns implicitly. 
Hence, it is reasonable to construct a classifier $C$ by utilizing the dynamic temporal pattern straightforwardly to detect FoG events along with the temporal index $t$. 
In detail, it contains fully connected (FC) layers with proper activation functions (LeakyReLU functions for all FC layers except for the last one which adopts a sigmoid function) to generate the predictions.
We denote $\hat{y}_{t}^{m, n}$ as the estimation of $y_{t}^{m, n}$. 

As FoG detection can be viewed as a binary classification problem, a binary cross entropy loss function was applied for the optimization purpose:
\begin{eqnarray}
\min_{C,G} J_{c} = -\frac{1}{MN}\sum_{m}^{M}\sum_{n}^{N} \frac{1}{T^{m, n}} \sum_{t}^{T^{m,n}}[y_{t}^{m, n}log(\hat{y}_{t}^{m,n}) \nonumber \\
             +(1-y_{t}^{m, n})log(1-\hat{y}_{t}^{m,n})],
\label{equ:loss_1}
\end{eqnarray}
which optimizes the parameters of $G$ and $C$.

In general, optimizing this loss function is able to obtain proper parameters to extract the spatio-temporal representation for FoG detection. 
However, this representation could contain subject specific information existing in the training set, which could cause the over-fitting risk and negatively impact the detection performance when predicting on the subjects not seen during the training. 
Considering the inter-subject variance is non-negligible compared with the variance between FoG and non-FoG cases, it is necessary to reduce the inter-subject variance among the representations and to derive subject-independent FoG representations.

\subsection{Multi-level Subject Discriminator}

In this study, a subject discriminator was proposed to distinguish whether two given spatio-temporal representations are generated from the same subject distribution or not.
It worked together with an adversarial training scheme to reduce the inter-subject variance of the spatio-temporal FoG representation and to treat the task as a fine-grained classification problem. 
%Hence, the representation contains less subject specified information and concentrates more on FoG patterns.

Basically, for two subjects $m'$ and $m''$ and their trials $n'$ and $n''$, the spatio-temporal representations can be summarized as $G(\{\mathbf{X}_{t, p}^{m', n'}\})=\{\mathbf{s}^{m', n'}_{t}, \dot{\mathbf{g}}_{t}^{m', n'}, \ddot{\mathbf{g}}_{t}^{m', n'}\}$ and $G(\{\mathbf{X}_{t, p}^{m'', n''}\})=\{\mathbf{s}^{m'',n''}_{t}, \dot{\mathbf{g}}_{t}^{m'',n''}, \ddot{\mathbf{g}}_{t}^{m'',n''}\}$, respectively.  The discriminator $D(G(\{\mathbf{X}_{t, p}^{m', n'}\}), G(\{\mathbf{X}_{t, p}^{m'', n''}\}))$ differentiates whether the two representations are from the same subject distribution ($m' = m''$) or not ($m' \neq m''$).

In detail, the second-order difference of the two representations are computed as illustrated in Eq. (\ref{equ:delta_s}) - (\ref{equ:delta_g2}), where $(\cdot)^{\circ 2}$ represents element-wise square operator. 
The second-order difference can be viewed as a special case of the bilinear computation, which has been proven as an efficient approach for fine-grained classification. Hence, it is expected that the second-order difference can be beneficial for generating subject-independent representation. 

Note that the computations take the spatial, intrinsic temporal and dynamic temporal representations into account together, although the temporal patterns implicitly contain the spatial pattern. 
The idea is similar to the consideration of using identity block in ResNet, which alleviates gradient vanishing issues and helps adversarial training to transfer and update gradients from the early layers.
\begin{equation}
\bigtriangleup \mathbf{s}^{m',n',m'',n''}_{t} = (\mathbf{s}^{m', n'}_{t} - \mathbf{s}^{m'',n''}_{t})^{\circ 2},
\label{equ:delta_s}
\end{equation}
\begin{equation}
\bigtriangleup \dot{\mathbf{g}}_{t}^{m',n',m'',n''} = (\dot{\mathbf{g}}_{t}^{m', n'} - \dot{\mathbf{g}}_{t}^{m'',n''})^{\circ 2},
\label{equ:delta_g1}
\end{equation}
\begin{equation}
\bigtriangleup \ddot{\mathbf{g}}_{t}^{m',n',m'',n''} = (\ddot{\mathbf{g}}_{t}^{m', n'} - \ddot{\mathbf{g}}_{t}^{m'',n''})^{\circ 2}.
\label{equ:delta_g2}
\end{equation}

By concatenating the three second-order difference vectors in Eq. (\ref{equ:delta_s}) - (\ref{equ:delta_g2}), a neural network based discriminator can be constructed, which consists of one FC layer and a sigmoid activation function for binary outputs (i.e., 0 for $m'=m''$ and 1 for $m'\neq m''$). 
This simple architecture later helps to efficiently refine the parameters of the spatio-temporal representation generator $G$ during the adversarial training. 
Eq. (\ref{equ:loss_d}) illustrates the binary cross entropy loss function for optimizing the discriminator. Note that the parameters of $G$ is frozen during the optimization of $J_{D}$:
\begin{equation}
\min_{D} J_{D}=-\sum_{m' = m''}log(D(G(\mathbf{X}_{t, p}^{m', n'}\}), G(\mathbf{X}_{t, p}^{m'', n''}\}))) \nonumber
\end{equation}
\begin{equation}
- \sum_{m' \neq m''} log(1 - D(G(\mathbf{X}_{t, p}^{m', n'}\}), G(\mathbf{X}_{t, p}^{m'', n''})).
\label{equ:loss_d}
\end{equation}

\begin{figure*}[htbp]
\begin{minipage}[t]{0.49\textwidth}
\centering
\includegraphics[width=\textwidth]{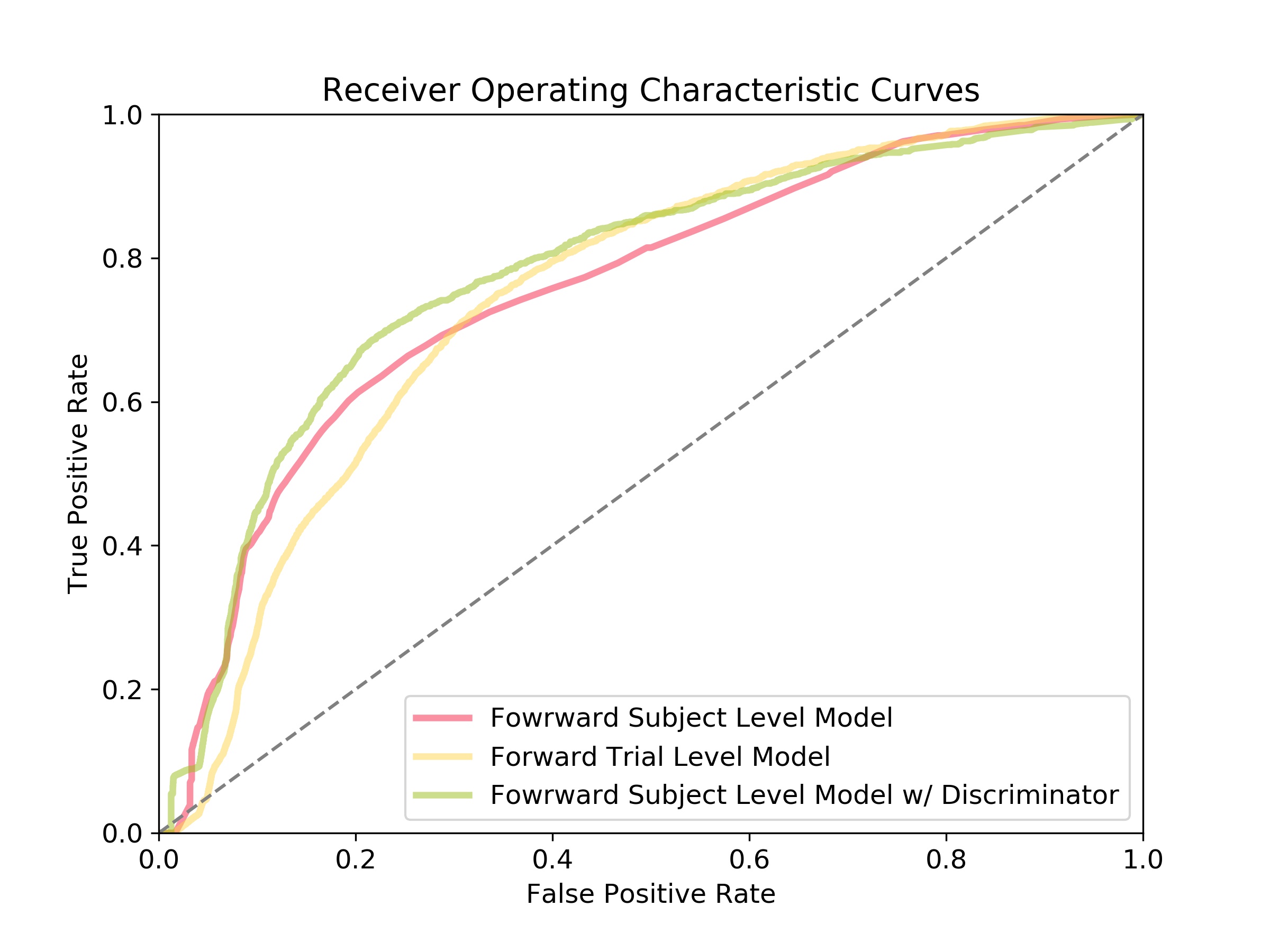}
\caption*{(a) }
\end{minipage}
\begin{minipage}[t]{0.49\textwidth}
\centering
\includegraphics[width=\textwidth]{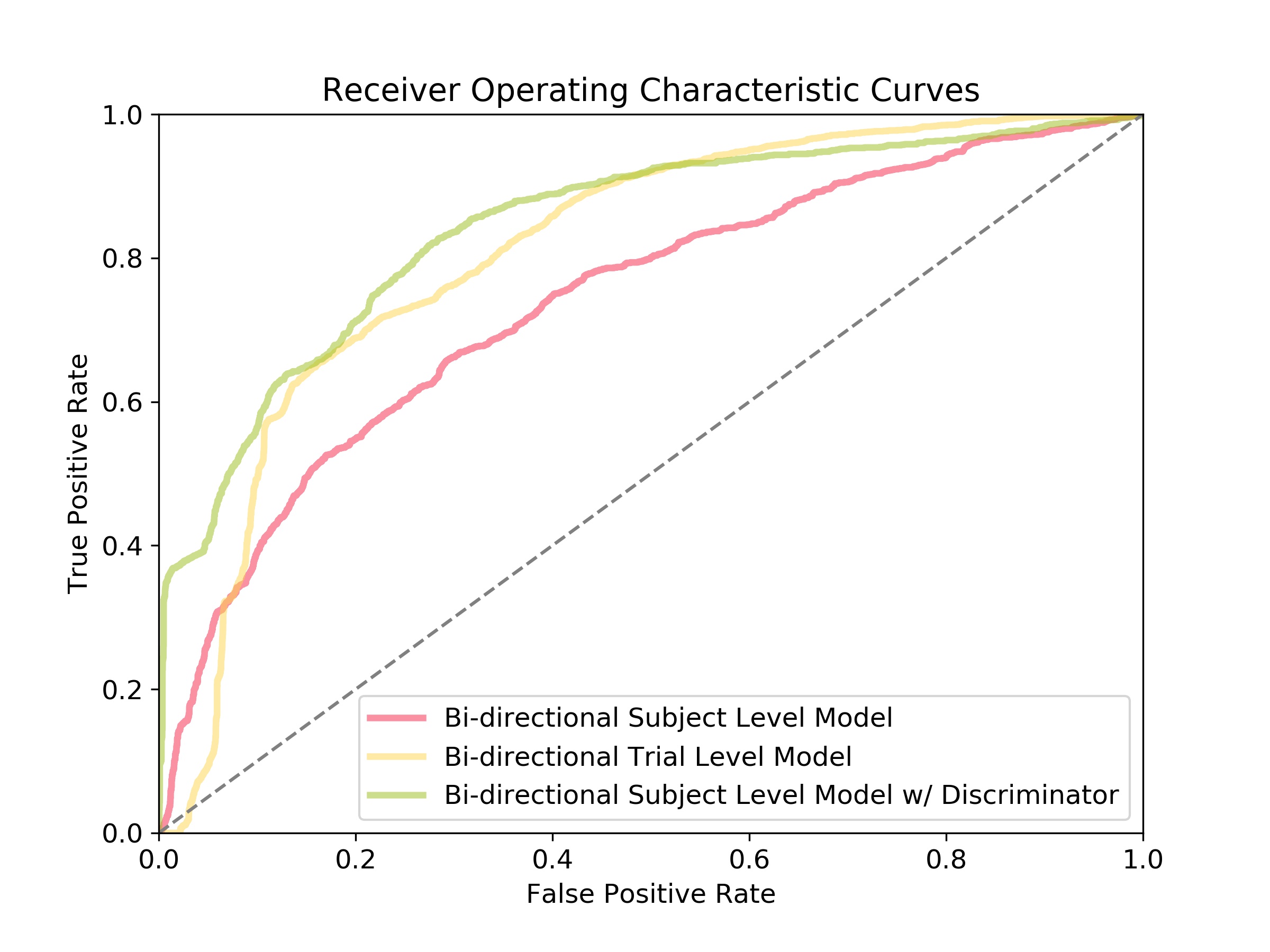}
\caption*{(b)}
\end{minipage}
\caption{ROC curves of the proposed methods. (a) Comparisons among the forward architectures with different training protocols. (b) Comparisons among the bi-directional architectures with different training protocols.}
\label{fig:ROC_GS_RNN}
\end{figure*}

\begin{center}
\begin{table*}[htbp]
  \centering
  \caption{FoG detection performance of our models in terms of different metrics}
    \begin{tabular}{cccccccccc}
    \toprule
          & AUC   & Youden's J & Sens. & Spec. & \tabincell{c}{False\\ positive\\ rate } & \tabincell{c}{False\\ negative\\ rate} & \tabincell{c}{Likelihood\\ ratio\\ positive} & \tabincell{c}{Likelihood\\ ratio\\ negative}  & Accuracy\\
    \midrule
    \multicolumn{10}{c}{Subject-level} \\
    \midrule
    Forward Model & 0.754 & 0.40  & 68.6\% & 71.7\% & 28.3\% & 31.4\% & 2.42  & 0.44  & 70.6\% \\
    Forward Model w/ Discirminator & 0.780 & 0.45  & 71.4\% & 73.2\% & 26.8\% & 28.6\% & 2.66  & 0.39  & 78.0\% \\
    Bi-directional Model & 0.798 & 0.47  & 76.5\% & 70.6\% & 29.4\% & 23.5\% & 2.60  & 0.33  & 71.8\% \\
    \textbf{Bi-directional Model w/ Discriminator} & \textbf{0.847} & \textbf{0.56} & \textbf{83.4\%} & \textbf{72.9\%} & \textbf{27.1\%} & \textbf{16.6\%} & \textbf{3.08} & \textbf{0.23} & \textbf{75.7\%} \\
    \midrule
    \multicolumn{10}{c}{Trial-level} \\
    \midrule
    Forward Model & 0.769 & 0.43  & 74.1\% & 68.5\% & 31.5\% & 25.9\% & 2.35  & 0.38  & 70.3\% \\
    Bi-directional Model & 0.819 & 0.50  & 73.9\% & 76.0\% & 24.0\% & 26.1\% & 3.08  & 0.34  & 75.2\% \\
    \bottomrule
    \end{tabular}%
  \label{tab:performance}%
\end{table*}%
\end{center}

\begin{figure*}[htbp]
\begin{minipage}[t]{0.49\textwidth}
\centering
\includegraphics[width=\textwidth]{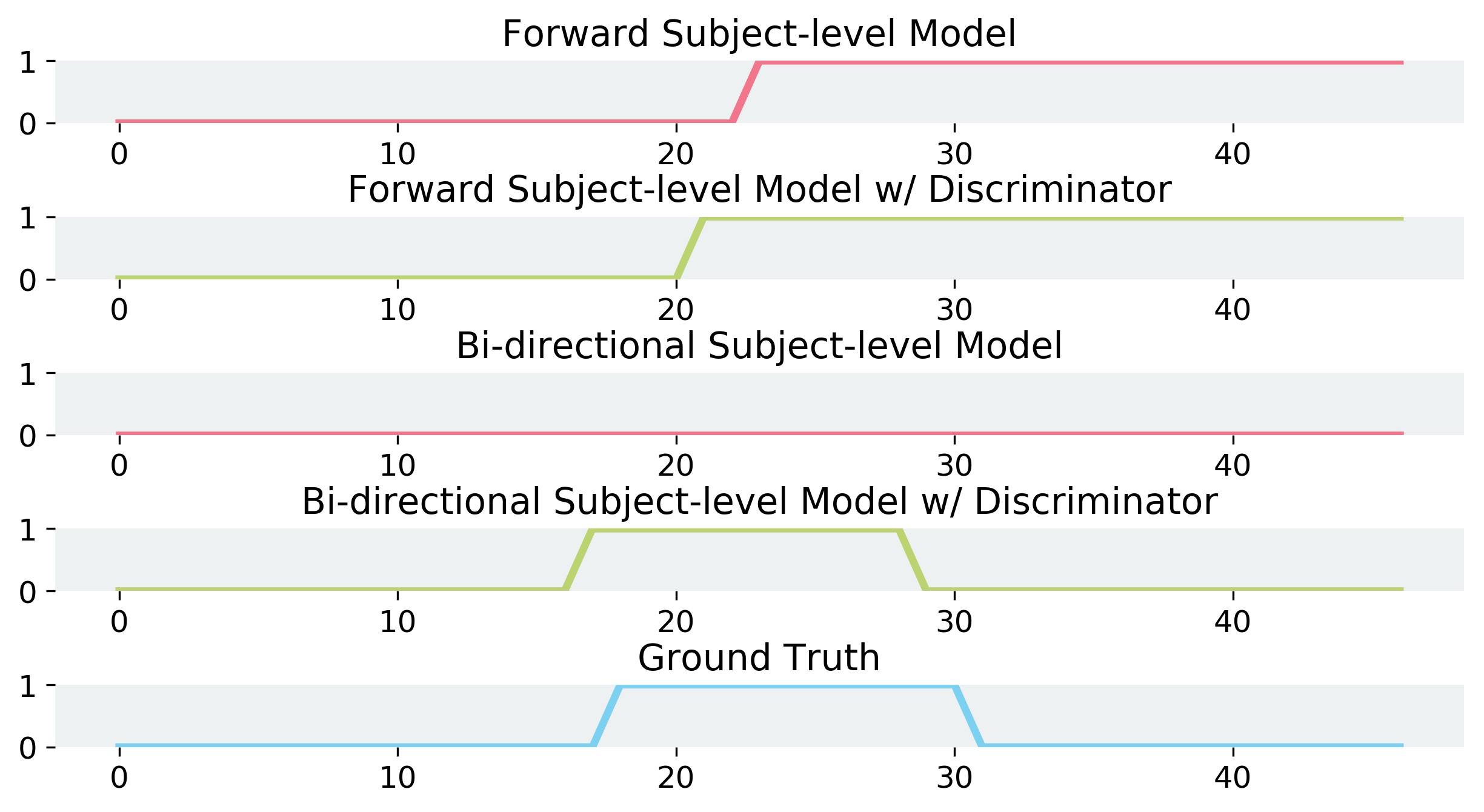}
\caption*{(a) }
\end{minipage}
\begin{minipage}[t]{0.49\textwidth}
\centering
\includegraphics[width=\textwidth]{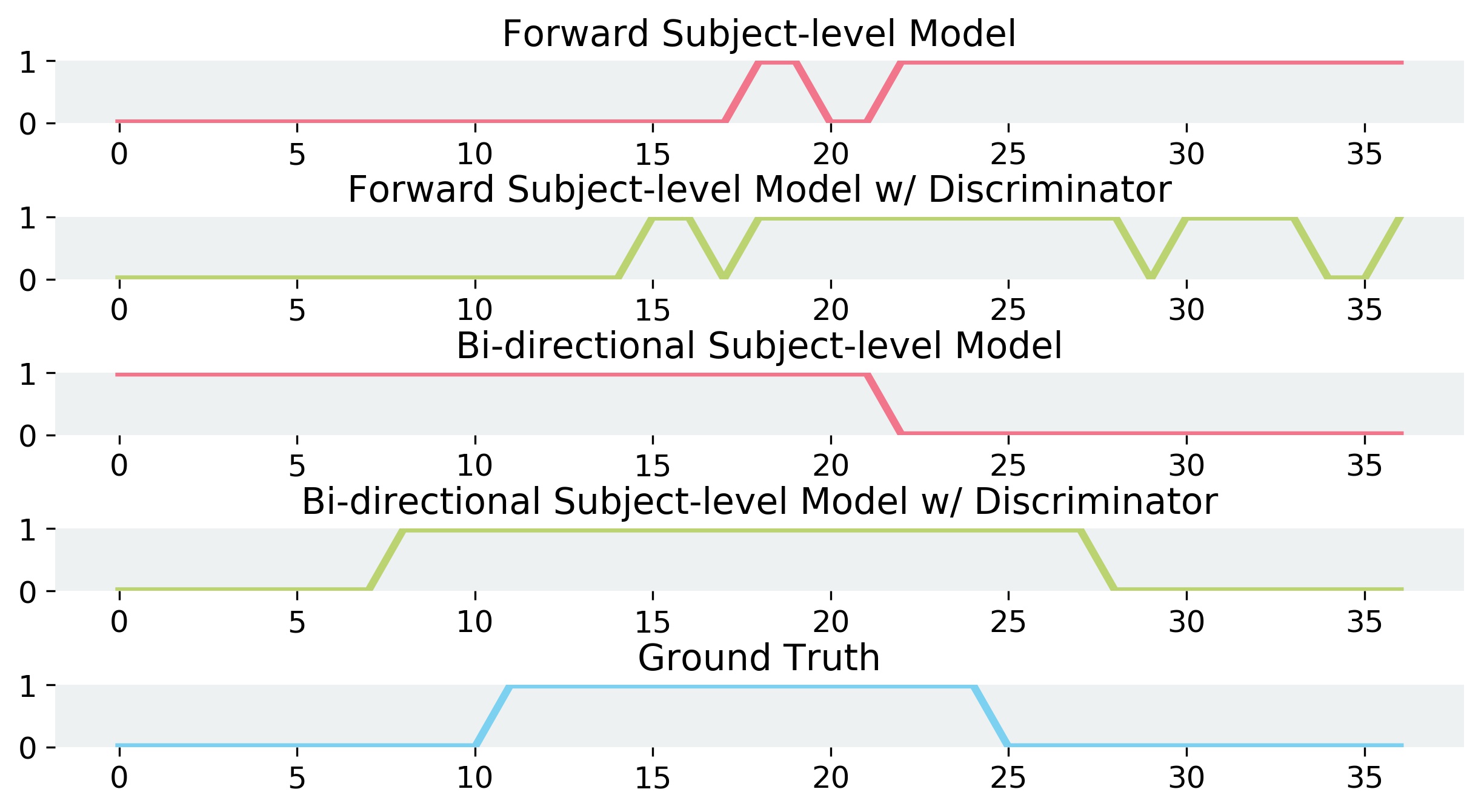}
\caption*{(b)}
\end{minipage}
\begin{minipage}[t]{0.49\textwidth}
\centering
\includegraphics[width=\textwidth]{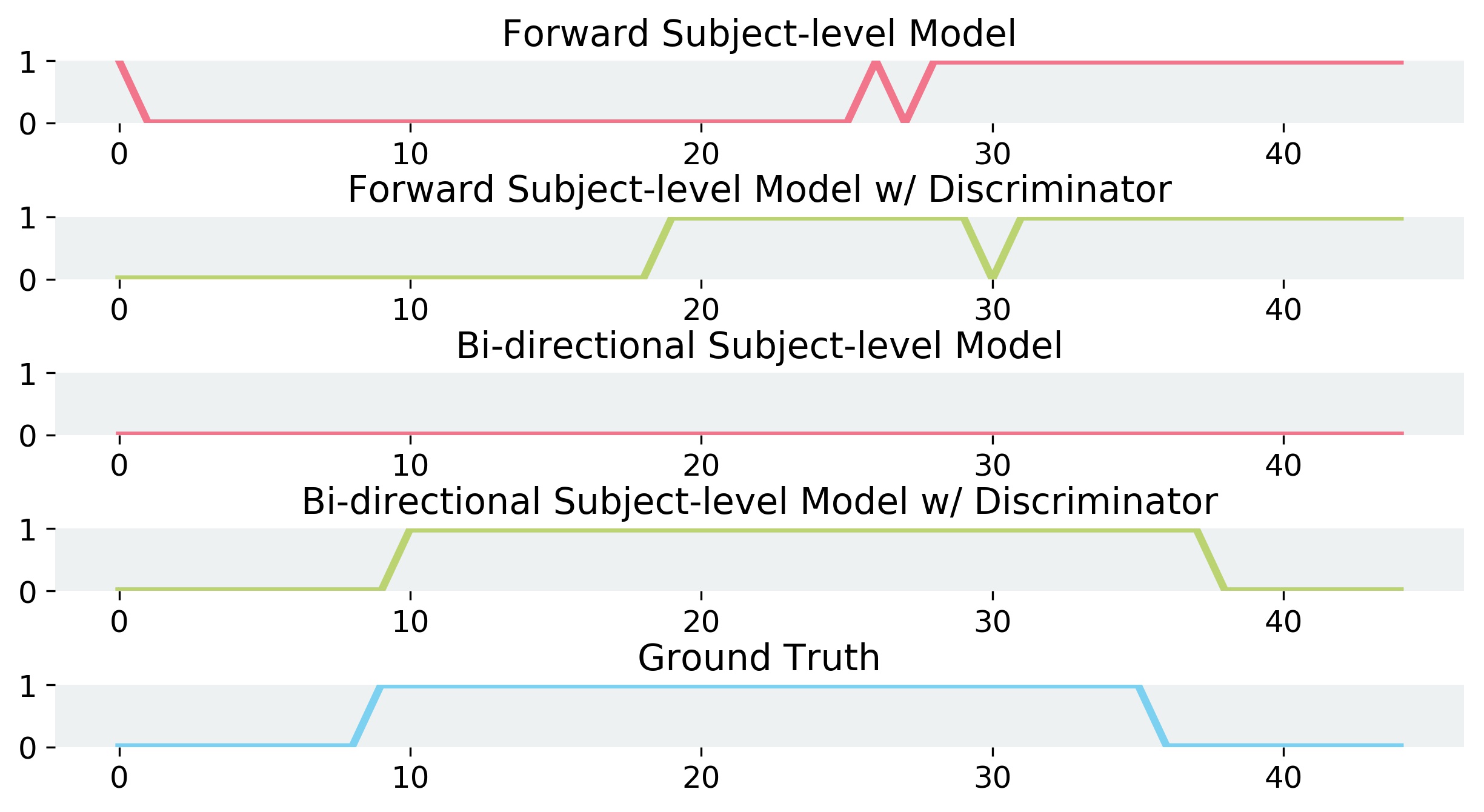}
\caption*{(b)}
\end{minipage}
\begin{minipage}[t]{0.49\textwidth}
\centering
\includegraphics[width=\textwidth]{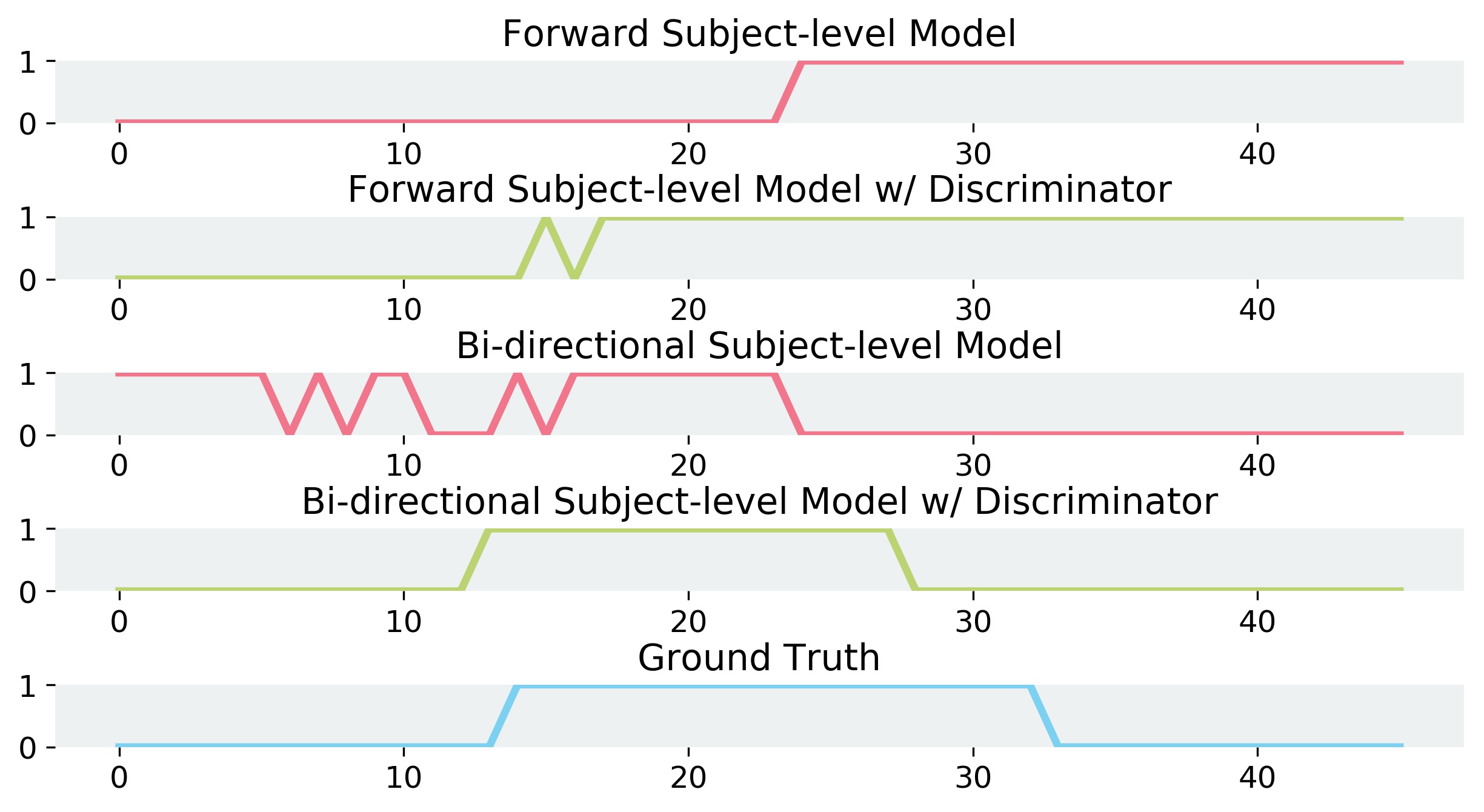}
\caption*{(b)}
\end{minipage}
\caption{Illustrations of the FoG detection results of four selected trials shown in (a) (b) (c) (d), which indicate the effects by introducing the adversarial training scheme. Compared with the ground truth data, both the forward and the bi-directional ASTN models characterize FoG events better than their counterparts without a subject discriminator involved.}
\label{fig:trailsamples}
\end{figure*}

\begin{algorithm}
\caption{End-to-end adversarial ASTN training}\label{euclid}
\hspace*{\algorithmicindent} \textbf{Input: } population of the footstep pressure sequences.\\
\hspace*{\algorithmicindent} \textbf{Output: } parameters of $G$, $C$ and $D$.
\begin{algorithmic}[1]
\While{not stopping criteria}
\State \Longunderstack[l]{sample four sequences for the current batch: \\ $\{\mathbf{X}_{t,p}^{m_{1},n_{1}}\}$, $\{\mathbf{X}_{t,p}^{m_{2},n_{2}}\}$, $\{\mathbf{X}_{t,p}^{m_{3},n_{3}}\}$ and $\{\mathbf{X}_{t,p}^{m_{3},n_{4}}\}$.}
\State \Longunderstack[l]{minimize $J_{C}$ (Eq. \ref{equ:loss_1}) and update the parameters of \\ $G$ and $C$ over the sampled sequences.}
\State \Longunderstack[l]{compute spatio-temporal representations \\ $G(\{\mathbf{X}_{t,p}^{m_{1},n_{1}}\})$, $G(\{\mathbf{X}_{t,p}^{m_{2},n_{2}}\})$, $G(\{\mathbf{X}_{t,p}^{m_{3},n_{3}}\})$ \\ and $G(\{\mathbf{X}_{t,p}^{m_{3},n_{4}}\})$ using the updated $G$.}
\State \Longunderstack[l]{minimize $J_{D}$ (Eq. \ref{equ:loss_d}) and update the parameters \\ of $D$ over the two pairs: $(G(\{\mathbf{X}_{t,p}^{m_{1},n_{1}}\})$, \\ $G(\{\mathbf{X}_{t,p}^{m_{2},n_{2}}\}))$ and $(G(\{\mathbf{X}_{t,p}^{m_{3},n_{3}}\})$, $G(\{\mathbf{X}_{t,p}^{m_{3},n_{3}}\}))$, \\ while the parameters of $G$ and $C$ are frozen.}
\State \Longunderstack[l]{maximize $J_{A}$ (Eq. \ref{equ:loss_a}) over the latter pair \\ $(G(\{\mathbf{X}_{t,p}^{m_{3},n_{3}}\})$ $G(\{\mathbf{X}_{t,p}^{m_{3},n_{4}}\}))$ to update the \\parameters of $G$, while keeping the parameters of\\ $D$ frozen.}
\EndWhile
\end{algorithmic}
\label{alg1}
\end{algorithm}

\subsection{Adversarial Training}

To let the FoG classifier and the subject discriminator work together, the adversarial training, which was originally proposed for generation tasks \cite{radford2015unsupervised}, was introduced so that the training was not limited to the two independent loss functions proposed as $J_{C}$ and $J_{D}$. %Adversarial training is widely used in generative models to capture the distribution of the training data. In general, it contains a generator to spawn fake data points to be similar to the training data and a discriminator to judge if the data points are from the training distribution or faked ones. The generator and discriminator are trained together to achieve the goal. Similarly, 
%To optimize the classifier, it generates spatio-temporal representation to characterize FoG patterns. To reduce subject variance, the subject discriminator judges if the representation are from the same subject or not.
During the adversarial training, the classifier continuously attempts to generate better FoG representations for improving the FoG classification performance related to the training data, whilst the subject discriminator was trained to become a better detective for correctly judging whether two representations are uniquely distributed or not.
The equilibrium of this game is achieved when the classifier is able to detect FoG patterns accurately, and the subject discriminator is left to always randomly guess at 50\% confidence for the representations.
In detail, the adversarial training optimizes the loss function $J_{A}$ in Eq. (\ref{equ:loss_a}), which combines Eq. (\ref{equ:loss_1}) and Eq. (\ref{equ:loss_d}) to play a minimax game.
\begin{equation}
\max_{G} J_{A}= - \sum_{m' \neq m''} log(1 - D(G(\mathbf{X}_{t, p}^{m', n'}), G(\mathbf{X}_{t, p}^{m'', n''})).
\label{equ:loss_a}
\end{equation}

Maximizing $J_{A}$ optimizes the parameters of $G$ to produce subject-independent spatio-temporal representations, which confuses the discriminator. Note that the parameters of the discriminator $D$ is frozen during the optimization of $J_{A}$.
With the above discussions, now an end-to-end adversarial training scheme for the footstep pressure sequence can be derived. Algorithm \ref{alg1} illustrates the key steps of our adversarial training scheme.

\section{Experimental Results and Discussions}
\label{sec:evaluation}

\begin{figure}[htbp]

\begin{minipage}[t]{0.49\textwidth}
\centering
\caption*{Subject-level split protocol}
\end{minipage}

\begin{minipage}[t]{0.24\textwidth}
\centering
\includegraphics[width=\textwidth]{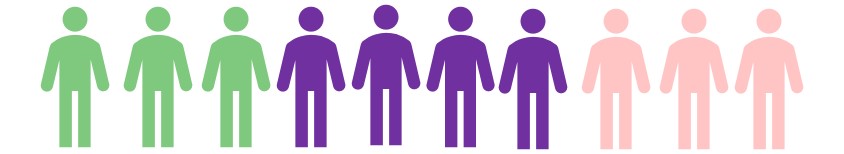}
\caption*{(a) training/val partition}
\end{minipage}
\begin{minipage}[t]{0.24\textwidth}
\centering
\includegraphics[width=\textwidth]{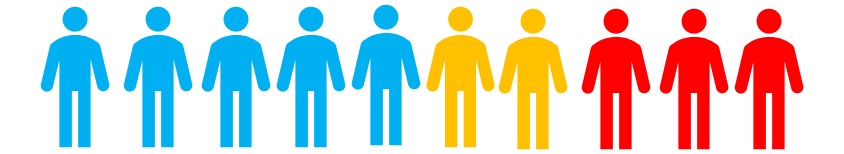}
\caption*{(b) test partition}
\end{minipage}

\begin{minipage}[t]{0.49\textwidth}
\centering
\caption*{\\Trial-level split protocol}
\end{minipage}

\begin{minipage}[t]{0.24\textwidth}
\centering
\includegraphics[width=\textwidth]{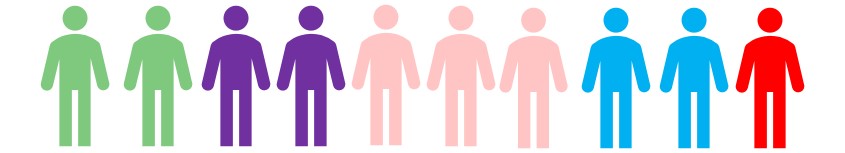}
\caption*{(c) training/val partition}
\end{minipage}
\begin{minipage}[t]{0.24\textwidth}
\centering
\includegraphics[width=\textwidth]{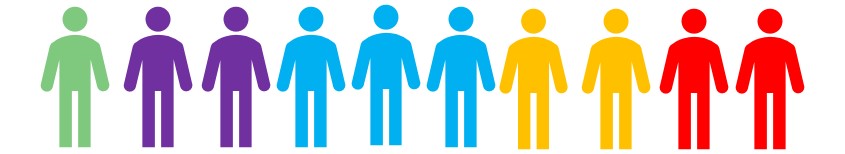}
\caption*{(d) test partition}
\end{minipage}

\caption{Illustration of the two protocols for the training-test split. Each colored shape represents a trail of which the color identifies one particular subject. 
For the subject-level split protocol, trials belonging to one subject either appear in the training partition or the test partition. For the trial-level split protocol, trials from one subject can appear in both the training and test partitions.}
\label{fig:splitting}
\end{figure}

\subsection{Dataset}

The FoG footstep pressure dataset used in this study consists of 21 subjects who participated in our clinical assessment. All the subjects were required to finish a number of TUG gait trials. In each trial, the subjects followed the instruction to walk within specified scenarios (i.e. either across an elevated plank that varied in width, or across a ground plank that varied in width) presented in virtual reality through an HTC VIVE head mounted display. 
In total, 393 trials were recorded by video and a Zeno pressure sensor gait mat (Protokinetics) simultaneously. Videos were collected for the purpose of FoG annotation by well-trained experts at the video frame level. 
To collect pressure sensing data, $500 \times 50$ sensors, which collected 10 levels of force response were placed in a Zeno gait mat, sampling at 120 Hz during the trial.
In total, 16,716 seconds of pressure signals were obtained.
For FoG detection, these pressure recordings were divided into 1-second long periods of non-overlapping temporal segments, and each segment was classified as either FoG or non-FOG class. 
If any frame of a segment was annotated as FoG frames according to the ground truth, the pressure data segment was labelled as FoG; otherwise, it was labelled as non-FoG.
Among these segments, 3,816 seconds contained FoG patterns confirmed by the video data and were marked as FoG events, while the rest of the trials were denoted as non-FoG events.
The event rate of this dataset is thus 22.8\%, which indicates that the dataset is imbalanced as one would expect given that FoG happens during the minority of the patient’s gait.

\subsection{Experimental Settings}

It is common that machine learning based clinical studies are devised in line with a number of trials collected from each subject. To optimize and evaluate these machine learning models, the trials are usually split into training and test partitions. Note that under this protocol the trials of one unique subject had the chance to appear in both the training and the test stages. 
Hence, it is possible to over-estimate the model performance considering that the subject-level predictor-response relations had been already leaked during the training stage especially for high subject-variance cases. 
Thus, it makes sense to follow a subject-level split manner, i.e., trials of one specified subject appear either in the training partition or in the test partition, which helps objectively estimate the effectiveness of the modelling for unseen subjects.
Fig. \ref{fig:splitting} illustrate the sample split protocols of the subject-level and the trial-level. 
%Therefore, the challenge is to adopt this subject-level split manner while achieving satisfying performance for FoG detection get rid of the impacts of the high subject-variance.

Therefore, in this study, three types of architectures are constructed to demonstrate the effectiveness of the proposed method. The details of the three architectures are introduced as follows:
\begin{itemize}
  \item \textbf{Subject-level model} is trained and tested following the subject split manner. The subject split protocol helps to verify the effectiveness of the modelling method for the assessment of unseen subjects. Note that this architecture does not contain the discriminator and it only optimizes the loss function $J_{C}$ defined in Eq. (\ref{equ:loss_1}).
  \item \textbf{Trial-level model} shares the same architecture as the subject-level model. However, it follows the trial split protocol. As subject-level related patterns could be already revealed during the training stage, it is expected that optimistic performance can be obtained compared to the subject-level model.
  \item \textbf{Subject-level model with discriminator (ASTN)} and adversarial training scheme is introduced to reduce the inter-subject variance of the spatio-temporal representation. This model is expected to be robust for unseen subjects and to gain better performance than the subject-level model.
\end{itemize}

Moreover, forward and bi-directional counterparts are implemented for all these three architectures by altering the direction of the GRU part. 
To comprehensively evaluate the performance of the proposed method, for each architecture, 5 random training-test data split are implemented and average performance metrics over these splits are computed as the estimated model performance. 
To help evaluate the subject-level models with sufficient subject samples, the ratio of the subject count in each training-test split is set to 1:1 for subject-level models. Similarly, the ratio of trial count in each split of the trial-level model is set to 1:1 as well. 
Note that a subset of the training data is withheld for validation purposes.
All models are trained from scratch and Adam optimizer is utilized with an initial learning rate of 0.001.

\subsection{FoG Detection Performance of ASTN}

A number of metrics are adopted to measure the FoG detection performance due to the imbalanced nature of the dataset. 
Firstly, as the prediction $\hat{y}_{t}^{m, n}$ is continuous in $[0, 1]$, which represents FoG probability, a threshold should be identified in line with the use cases. Next, accuracy (the count of samples correctly classified over the total sample size), sensitivity (true positive rate) and specificity (true negative rate) associated with the threshold can be computed.
By varying the threshold and plotting the corresponding sensitivity against 1-specificity, receiver operating characteristic (ROC) curve and the area under (ROC) curve (AUC) are further utilized to evaluate the diagnosis ability of the proposed ASTN architecture. 

To show the effectiveness of the proposed method, these metrics are evaluated under the three architectures including subject-level model, trial-level model and subject-level model with a discriminator (ASTN) trained in an adversarial manner. 
Forward and bi-directional GRUs are implemented for all these architectures: the forward direction only involves the historical footstep pressure maps and could be utilized for online prediction; the bi-directional networks utilize both the historical and future footstep pressure maps to obtain more accurate spatio-temporal representations.
Fig. \ref{fig:ROC_GS_RNN} (a) shows the ROC curves for comparison purposes of these methods and 
Table \ref{tab:performance}
lists the AUC values of these ROC curves. 
In summary, bi-directional subject-level with discriminator model achieves the best performance in terms of AUC of 0.847 compared to the other architectures. Note that all bi-directional models outperform their forward counterparts.

\kun{As expected, for both the forward and the bi-directional architectures, the performances of the subject-level models drop compared to those of the trial-level models in terms of AUC: 0.754 for the forward subject-level model vs. 0.769 for the forward trial-level model; 0.798 for the bi-directional subject-level model vs. 0.819 for the bi-directional trial-level model. These results confirm the assumptions that subject-level patterns could lead to an over-fitting issue in a trial-level model, in which a subject can appear in both training and test splits. Such settings can inflate the performance and reduce the reliability for unseen subjects who possess different subject-level patterns. The proposed discriminator and adversarial training strategy help to alleviate this issue and increase the performances of the subject-level models with all kinds of GRU architectures (forward and bi-directional). Particularly, with the adversarial training strategy, the performances of the subject-level models achieve superior performance even compared to the inflated performances of the trial-level models.}

In particular, the sensitivity, specificity and accuracy related to a threshold $\hat{\theta}$ are also adopted for evaluation, which maximizes the following Youden's J statistics \cite{youden1950index}:
$$\hat{\theta} = \argmin_\theta sensitivity + specificity - 1.$$
By maximizing the statistic, a threshold can be derived to treat sensitivity and specificity with equal importance.
The evaluation metrics and associated J statistics are also listed in Table. \ref{tab:performance}.
For the best model in terms of AUC, sensitivity, specificity and accuracy values related to this threshold achieve 83.4\%, 72.9\% and 75.7\%, respectively.

As illustrated in Fig. \ref{fig:trailsamples}, the FoG detection results of selected trials demonstrate the effectiveness of introducing the adversarial training scheme. Compared to the ground truth data, both the forward and the bi-directional ASTNs characterize FoG events better than their counterparts without involving the discriminator.

\begin{figure}[!htb]
\centering
\includegraphics[width=0.49\textwidth]{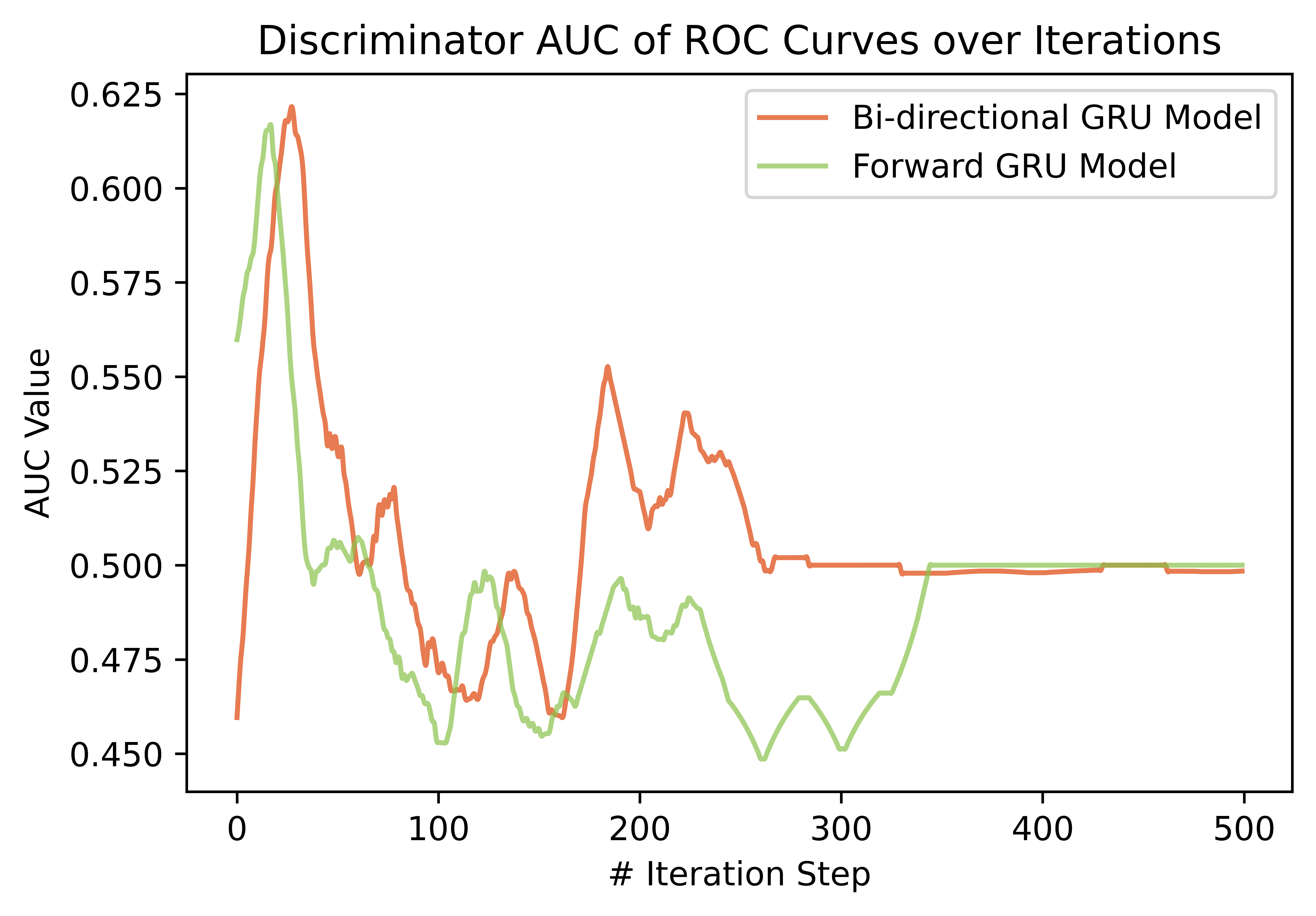}
\caption{\kun{Illustration of the performance of the discriminator for the bi-directional and the forward ASTN architectures. AUC values against iteration steps are shown to indicate the advantage of the discriminator. }}
\label{fig:discriminator}
\end{figure}

\begin{figure*}[!htb]
\centering
\includegraphics[width=1\textwidth]{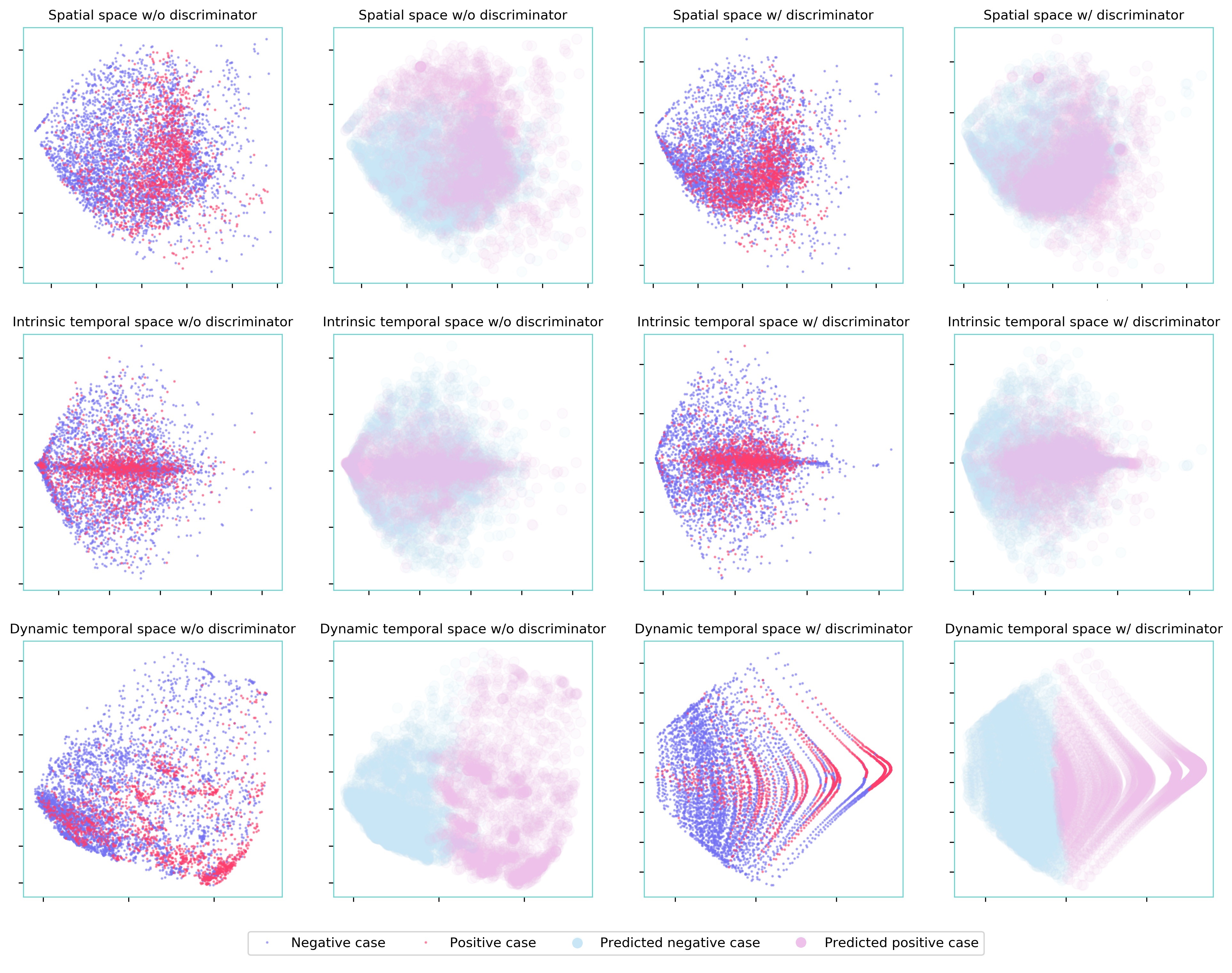}
\caption{Illustration of the feature vector distribution. The first two components of the representation at each level are selected and drawn in a 2-dimensional space. For each level, the sub-figures are associated with and without using the proposed adversarial training scheme. In particular, we use two sub-figures for each case to clearly present the ground truth and the predicted labels in line with the distribution, respectively.}
\label{fig:key_vertices}
\end{figure*}

\subsection{Impact of Discriminator}

The discriminator determines whether any two representations belong to a unique subject distribution or not. 
It is expected that it would be increasingly challenging to distinguish the representations from distinct subject distributions along the adversarial training. 
Fig. \ref{fig:discriminator} shows the AUC values of the discriminator ROC curves against the iteration steps. 
Along the iteration, both the discriminators of the forward and the bi-directional ASTNs tend to achieve an AUC around 0.5, which indicate that it is approximately random in discriminating subject distributions of the input pair with their spatio-temporal representations. 
It further implies that the inter-subject variation of the spatio-temporal representation has been significantly reduced, which could be beneficial for the generalization of the classifier for undertaking FoG detection on unseen subjects. 
In addition, it can be observed that the bi-directional architecture converges with fewer iterations than the forward counterpart. These findings suggest that the adversarial optimization of both forward and backward patterns increase the effectiveness to eliminate the subject associated redundancy. 

\begin{table}[htbp]
  \centering
  \caption{Comparison of discriminator architectures: different constructions of difference feature vectors}
    \begin{tabular}{p{7cm}c}
    
    \toprule
    Discriminator type   & AUC \\
    \midrule
    \multicolumn{2}{l}{Forward ASTN architecture} \\
    \midrule
    First-order difference discriminator & 75.6 \\
    Absolute first-order difference discriminator & 0.776 \\
    \textbf{Second-order difference discriminator} & \textbf{0.780} \\
    Concatenated discriminator & 0.760 \\
    \midrule
    \multicolumn{2}{l}{Bi-directional ASTN architecture} \\
    \midrule
    First-order difference discriminator & 0.800 \\
    Absolute first-order difference discriminator & 0.825 \\
    \textbf{Second-order difference discriminator} & \textbf{0.847} \\
    Concatenated discriminator & 0.846 \\
    \bottomrule
    \end{tabular}%
  \label{tab:discriminator_selection}%
\end{table}%

Table \ref{tab:discriminator_selection} further presents the performance comparisons of different discriminator architectures. In addition to the second-order difference discriminator, the first-order difference discriminator, the absolute first-order difference discriminator and the general concatenated discriminator are involved for the comparison. It can be observed that the proposed second-order difference discriminator achieves the best performance in terms of AUC for both the forward and the bi-directional ASTNs. 
Note that most of these discriminators outperform the corresponding trial-level models, which indicates the necessity and effectiveness to integrate the discriminator with an adversarial training scheme for subject-level clinical studies.

Note that the discriminator adopts spatial, intrinsic temporal and dynamic temporal representations simultaneously, which helps to transfer and update gradients within the layers near the input layer. This multi-level representation design plays a similar role to the mechanism of ResNet and DenseNet. 
To verify this design, we implement the counterparts for both forward and bi-directional ASTNs with discriminators only involving the dynamic temporal representations, which is close to the output layer instead of the input layer. 
As shown in Table \ref{tab:discriminator_mech} for the performance of these two strategies, the discriminator based on the multi-level representation outperform the one based on single dynamic temporal representation in terms of AUC. 

\begin{table}[htbp]
  \centering
  \caption{Comparison of discriminator architectures: multi-level vs. output-level only discriminators}
    \begin{tabular}{p{7cm}c}
    \toprule
    Discriminator type   & AUC \\
    \midrule
    \multicolumn{2}{l}{Forward ASTN architecture} \\
    \midrule
    \textbf{Multi-level representation} & \textbf{0.780} \\
    Dynamic temporal representation only & 0.771 \\
    \midrule
    \multicolumn{2}{l}{Bi-directional ASTN architecture} \\
    \midrule
    \textbf{Multi-level representation} & \textbf{0.847} \\
    Dynamic temporal representation only & 0.795 \\
    \bottomrule
    \end{tabular}%
  \label{tab:discriminator_mech}%
\end{table}%

\begin{figure}[!htb]
\centering
\includegraphics[width=0.49\textwidth]{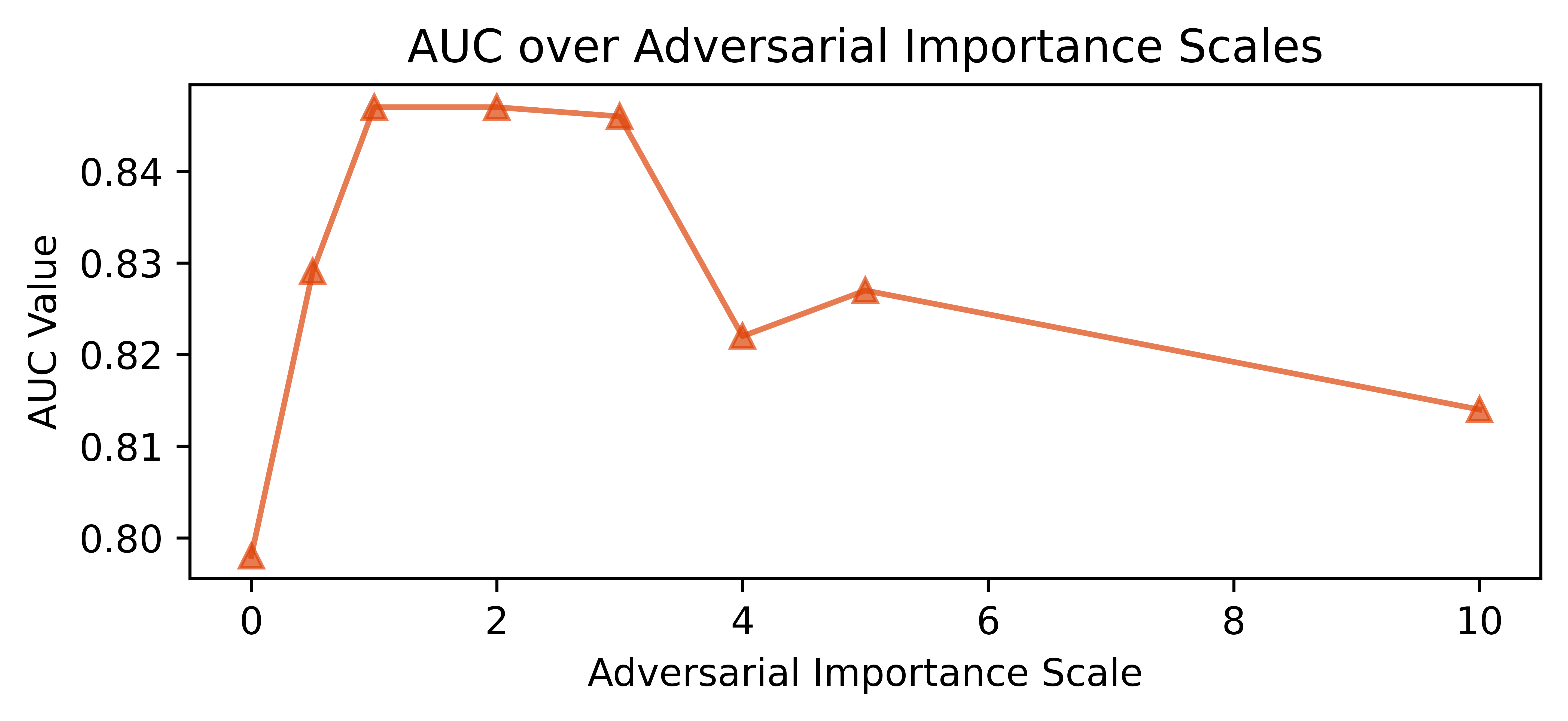}
\caption{Illustration of the performance of the bi-directional model with different adversarial scales. AUC values against adversarial scales are shown to indicate the impacts of the adversarial learning. }
\label{fig:auc_over_adversarial_scale}
\end{figure}

\kun{Moreover, Fig. \ref{fig:auc_over_adversarial_scale} illustrates the AUC of the bi-directional GRU model against different importance settings of the discriminator during the adversarial training. By default, the importance of the losses defined in Eq. (\ref{equ:loss_d}) and Eq. (\ref{equ:loss_a}) is set to 1 as in Algorithm 1. Particularly, the importance can be viewed as 0 for a model without adversarial training. It can be observed that the performance increases first with the increasing importance of the adversarial training and then decreases when the importance is too strong. The results suggest that the default setting is a reasonable choice for the FoG detection task. }

\subsection{Analysis of Representation Distribution}

To further understand the improvements of the representation space regularized by the adversarial training scheme with the multi-level subject discriminator, principal components analysis (PCA) is conducted on the three levels of representations individually, including the spatial representation, the intrinsic temporal representation and the dynamic temporal representation. 
As shown in Fig. \ref{fig:key_vertices}, each row illustrates the representation at a specific level with and without the adversarial training scheme. For each training scheme, the first two components of the PCA results are visualized in a 2-dimensional space. To clearly present the PCA results, we use different markers for the data points in line with their ground truth labels and predicted labels:
the red and the blue point markers denote the ground truth labels of the positive cases and the negative cases, respectively; the red and the blue circle markers denote the predicted positive cases and the predicted negative cases, respectively.

For the spatial representation space, the model with a discriminator tends to generate more separable data distribution for the two classes, whilst the model without a discriminator showed dispersed data points of each class over an extensive area. 
In terms of the predicted results, the separable data distribution helps to derive more discriminant regions to distinguish the positive cases from the negative cases.
In contrast, without the discriminator, the regions for the two classes tend to be undifferentiated, which is the cause of the decreased performance of the subject-level model. 
Similar results can be observed from the intrinsic and dynamic temporal representation spaces. 
Therefore, the adversarial training scheme with the multi-level subject discriminator contributed to the robustness of the representation space, and thus the model performance was superior to the trial-level model and the subject-level model only.

\section{Conclusion}
\label{sec:conclusion}

To investigate footstep pressure based FoG detection for the first time, this paper presents a novel architecture ASTN to process the data modality of footstep pressure sequences. 
ASTN involves a classifier to learn the spatio-temporal FoG representation from pressure sequences and a subject discriminator to regularize the representation to be subject-independent using the adversarial training procedure. 
The subject-independent characteristic is critical for clinical studies to ensure the effectiveness of the modelling methods to be deployed for unseen subjects. 
Note that this adversarial training scheme for the subject-independent representation can be integrated with many existing neural network architectures, and thus may improve other subject-level clinical studies.
Experimental results on a large in-house dataset highlights the superior performance of ASTNs compared to those analytic counterparts not exploiting the subject independent FoG representation.
Our future work will improve the discriminator architecture and enhance the FoG detection performance. It would also be helpful to apply the multi-level adversarial scheme to multi-modal FoG data to generate subject-independent cross-domain representations for more accurate characterization of FoG patterns.
%In addition, it is interesting that the ability of the architecture for fine-tuning an existing model to directly decrease the distribution gap and applied for new subjects.

\section*{Acknowledgement}
We thank our patients who participated into the data collection and they all provided written informed consent. The ethical approval of this research was obtained from the University of Sydney Human Ethics Board (\#2014/255). 
We would like to acknowledge and thank Kristin Economou for conducting and scoring the freezing of gait assessments. 

\bibliographystyle{ieeetr}
\bibliography{main}

\end{document}